\documentclass[11pt]{article}
\usepackage{eacl2012}
\usepackage{times}
\usepackage{latexsym}
\usepackage{amsmath}
\usepackage{cyrillic}
\usepackage{multirow}
\usepackage{url}

\newcommand{\preslav}[1]{}
\newcommand{\joro}[1]{}

\title{Feature-Rich Part-of-speech Tagging\\
       for Morphologically Complex Languages: Application to Bulgarian\\}

\author{\\\bf Georgi Georgiev and Valentin Zhikov\\
  Ontotext AD\\
  135 Tsarigradsko Sh., Sofia, Bulgaria\\
  {\tt \scriptsize{\{georgi.georgiev,valentin.zhikov\}@ontotext.com}} \\\And
  \\\bf Petya Osenova and Kiril Simov\\
  IICT, Bulgarian Academy of Sciences\\
  25A Acad. G. Bonchev, Sofia, Bulgaria\\
  {\tt \small{\{petya,kivs\}@bultreebank.org}} \\\AND
  Preslav Nakov \\
  Qatar Computing Research Institute, Qatar Foundation\\
  Tornado Tower, floor 10, P.O. Box 5825, Doha, Qatar \\
  {\tt \small{pnakov@qf.org.qa}} \\}

\date{}

\begin{document}
\maketitle
\begin{abstract}
    We present experiments with part-of-speech tagging for Bulgarian,
    a Slavic language with rich inflectional and derivational morphology.
    Unlike most previous work,
    which has used a small number of grammatical categories,
    we work with
    680 morpho-syntactic tags.
    We combine a large morphological lexicon with prior linguistic knowledge and guided learning
    from a POS-annotated corpus, achieving accuracy of 97.98\%,
    which is a significant improvement over the state-of-the-art for Bulgarian.
\end{abstract}

\section{Introduction}
\label{sec:intro}

Part-of-speech (POS) tagging is the task of assigning each of the words in a given piece of text
a contextually suitable grammatical category.
This is not trivial since words can play different syntactic roles in different contexts,
e.g., \emph{can} is a noun in ``\emph{I opened a can of coke.}''
but a verb in ``\emph{I can write.}''
Traditionally, linguists have classified English words into the following eight basic POS categories:
noun, pronoun, adjective, verb, adverb, preposition, conjunction, and interjection;
this list is often extended a bit, e.g., with determiners, particles, participles, etc.,
but the number of categories considered is rarely more than 15.

Computational linguistics works with a larger inventory of POS tags,
e.g., the Penn Treebank \cite{Marcus:1993:penntreebank} uses 48 tags:
36 for part-of-speech, and 12 for punctuation and currency symbols.
This increase in the number of tags is partially due to finer granularity,
e.g., there are special tags for determiners, particles, modal verbs, cardinal numbers, foreign words, existential \emph{there}, etc.,
but also to the desire to encode morphological information
as part of the tags.

For example, there are six tags for verbs in the Penn Treebank:
VB  (verb, base form; e.g., \emph{sing}),
VBD (verb, past tense; e.g., \emph{sang}),
VBG (verb, gerund or present participle; e.g., \emph{singing}),
VBN	(verb, past participle; e.g., \emph{sung})
VBP (verb, non-3rd person singular present; e.g., \emph{sing}), and
VBZ (verb, 3rd person singular present; e.g., \emph{sings});
these tags are morpho-syntactic in nature.
Other corpora have used even larger tagsets,
e.g., the Brown corpus \cite{kucera:computational:1967} and the Lancaster-Oslo/Bergen (LOB) corpus \cite{LOB}
use 87 and 135 tags, respectively.

POS tagging poses major challenges for morphologically complex languages,
whose tagsets encode a lot of additional morpho-syntactic features (for most of the basic POS categories),
e.g., gender, number, person, etc.
For example, the BulTreeBank~\cite{bul-morph} for Bulgarian uses 680 tags,
while the Prague Dependency Treebank~\cite{hajic:1998} for Czech has over 1,400 tags.

Below we present experiments with POS tagging for Bulgarian,
which is an inflectional language with rich morphology.
Unlike most previous work, which has used a reduced set of POS tags,
we use all 680 tags in the BulTreeBank.
We combine prior linguistic knowledge and statistical learning,
achieving accuracy comparable to that reported for state-of-the-art systems for English.

The remainder of the paper is organized as follows:
Section~\ref{sec:related} provides an overview of related work,
Section~\ref{sec:bulgarian} describes Bulgarian morphology,
Section~\ref{sec:method} introduces our approach,
Section~\ref{sec:dataset} describes the datasets,
Section~\ref{sec:experiments} presents our experiments in detail,
Section~\ref{sec:discussion} discusses the results,
Section~\ref{sec:analysis} offers application-specific error analysis,
and Section~\ref{sec:conclusion} concludes and points to some promising directions for future work.

\section{Related Work}
\label{sec:related}

Most research on part-of-speech tagging has focused on English,
and has relied on the Penn Treebank~\cite{Marcus:1993:penntreebank} and its tagset for training and evaluation.
The task is typically addressed as a sequential tagging problem; one notable exception
is the work of \newcite{Brill:1995:TEL}, who proposed non-sequential transformation-based learning.

A number of different sequential learning frameworks have been tried, yielding 96-97\% accuracy:
\newcite{Lafferty:2001:CRF} experimented with conditional random fields (CRFs) (95.7\% accuracy),
\newcite{Ratnaparkhi1996} used a maximum entropy sequence classifier (96.6\% accuracy),
\newcite{Brants00tnt} employed a hidden Markov model (96.6\% accuracy),
\newcite{Collins:2002:DTM} adopted an averaged perception discriminative sequence model (97.1\% accuracy).
All these models fix the order of inference from left to right.

\newcite{Toutanova:2003:FPT} introduced a cyclic dependency network (97.2\% accuracy),
where the search is bi-directional.
\newcite{shen-satta-joshi:2007:ACLMain} have further shown that better results (97.3\% accuracy) can be obtained using
guided learning,
a framework for bidirectional sequence classification,
which integrates token classification and inference order selection into a single learning task
and uses a perceptron-like \cite{collins-roark:2004:ACL} passive-aggressive classifier to make the easiest decisions first.
Recently, \newcite{tsuruoka-miyao-kazama:2011:CoNLL},
proposed a simple perceptron-based classifier applied from left to right
but augmented with a lookahead mechanism that searches the space of future actions,
yielding 97.3\% accuracy.

For morphologically complex languages,
the problem of POS tagging typically includes morphological disambiguation,
which yields a much larger number of tags.
For example, for Arabic, \newcite{Habash:2005:ATP} used support vector machines (SVM),
achieving 97.6\% accuracy with 139 tags from the Arabic Treebank~\cite{Arabic:treebank}.
For Czech, \newcite{hajickkop:2001} combined a hidden Markov model (HMM) with linguistic rules,
which yielded 95.2\% accuracy using an inventory of over 1,400 tags from the Prague Dependency Treebank~\cite{hajic:1998}.
For Icelandic, \newcite{dredze-wallenberg:2008:ACLShort} reported 92.1\% accuracy
with 639 tags developed for the Icelandic frequency lexicon~\cite{Pind:Icelandic:dict},
they used guided learning and tag decomposition:\\
First, a coarse POS class is assigned (e.g.,~noun, verb, adjective),
then, additional fine-grained morphological features like case, number and gender are added,
and finally, the proposed tags are further reconsidered using non-local features.
Similarly, \newcite{smith-smith-tromble:2005:HLTEMNLP} decomposed the complex tags into factors,
where models for predicting part-of-speech, gender, number, case, and lemma are estimated separately,
and then composed into a single CRF model;
this yielded competitive results for Arabic, Korean, and Czech.

Most previous work on Bulgarian POS tagging has started with large tagsets,
which were then reduced.
For example, \newcite{DBLP:conf/aimsa/DoychinovaM04} mapped their initial tagset of 946 tags to just 40,
which allowed them to achieve 95.5\% accuracy using the transformation-based learning of \newcite{Brill:1995:TEL},
and 98.4\% accuracy using manually crafted linguistic rules.
Similarly, \newcite{Georgiev:2009:CAB}, who used maximum entropy and the BulTreeBank~\cite{bul-morph},
grouped its 680 fine-grained POS tags into 95 coarse-grained ones,
and thus improved their accuracy from 90.34\% to 94.4\%.
\newcite{Simov:2001} used a recurrent neural network to predict
(a) 160 morpho-syntactic tags (92.9\% accuracy)
and (b) 15 POS tags (95.2\% accuracy).

Some researchers did not reduce the tagset:
\newcite{Savkov:al:2011} used 680 tags (94.7\% accuracy),
and \newcite{Tanev:Mitkov:2002} used 303 tags
and the BULMORPH morphological analyzer \cite{Krushkov},
achieving P=R=95\%.

\section{Bulgarian Morphology}
\label{sec:bulgarian}

Bulgarian is an Indo-European language from the Slavic language group, written with the Cyrillic alphabet
and spoken by about 9-12 million people.
It is also a member of the Balkan Sprachbund and thus differs from most other Slavic languages:
it has no case declensions,
uses a suffixed definite article (which has a short and a long form for singular masculine),
and lacks verb infinitive forms.
It further uses special evidential verb forms to express unwitnessed, retold, and doubtful activities.

Bulgarian is an inflective language with very rich morphology.
For example, Bulgarian verbs have 52 synthetic wordforms on average,
while pronouns have altogether more than ten grammatical features
(not necessarily shared by all pronouns), including case, gender, person, number, definiteness, etc.

This rich morphology inevitably leads to ambiguity proliferation;
our analysis of BulTreeBank shows four major types of ambiguity:

\begin{enumerate}
\item Between the wordforms of the same lexeme, i.e., in the paradigm.
     For example, \textcyrrm{divana}, an inflected form of \textcyrrm{divan} (`sofa', masculine),
     can mean (a) `the sofa' (definite, singular, short definite article)
     or (b) a count form, e.g., as in \textcyrrm{dva divana} (`two sofas').
\item Between two or more lexemes, i.e., conversion.
     For example, \textcyrrm{kato} can be (a) a subordinator meaning `as, when',
     or (b) a preposition meaning `like, such as'.
\item Between a lexeme and an inflected wordform of another lexeme, i.e., across-paradigms.
     For example, \textcyrrm{politika} can mean (a) `the politician' (masculine, singular, definite, short definite article)
     or (b) `politics' (feminine, singular, indefinite).
\item Between the wordforms of two or more lexemes, i.e., across-paradigms and quasi-conversion.
     For example, \textcyrrm{v\cdprime rvi} can mean (a) `walks' (verb, 2nd or 3rd person, present tense)
     or (b) `strings, laces' (feminine, plural, indefinite).
\end{enumerate}

Some morpho-syntactic ambiguities in Bulgarian are occasional, but many are systematic,
e.g., neuter singular adjectives have the same forms as adverbs.
Overall, most ambiguities are local, and thus arguably resolvable using $n$-grams,
e.g., compare \textcyrrm{hubavo dete} (`beautiful child'), where \textcyrrm{hubavo} is a neuter adjective,
and ``\textcyrrm{Peya hubavo.}'' (`I sing beautifully.'), where it is an adverb of manner.
Other ambiguities, however, are non-local and may require discourse-level analysis, e.g.,
``\textcyrrm{Vidyah go}.'' can mean `I saw him.', where \textcyrrm{go} is a masculine pronoun,
or 'I saw it.', where it is a neuter pronoun.
Finally, there are ambiguities that are very hard or even impossible\footnote{
The problem also exists for English,
e.g., the annotators of the Penn Treebank
were allowed to use tag combinations for inherently ambiguous cases:
JJ$|$NN (adjective or noun as prenominal modifier),
JJ$|$VBG (adjective or gerund/present participle),
JJ$|$VBN (adjective or past participle),
NN$|$VBG (noun or gerund),
and RB$|$RP (adverb or particle).}
to resolve,
e.g., ``\textcyrrm{Deteto vleze veselo.}'' can mean both
`The child came in happy.' (\textcyrrm{veselo} is an adjective)
and `The child came in happily.'  (it is an adverb);
however, the latter is much more likely.

In many cases, strong domain preferences exist about how various systematic ambiguities should be resolved.
We made a study for the newswire domain, analyzing a corpus of 546,029 words,
and we found that ambiguity type 2 (lexeme-lexeme) prevailed for functional parts-of-speech,
while the other types were more frequent for inflecting parts-of-speech.
Below we show the most frequent types of morpho-syntactic ambiguities and their frequency in our corpus:

\vspace{-1mm}
\begin{itemize}\itemsep0pt\parsep0pt\parskip0pt
    \item \textcyrrm{na}: preposition (`of') vs. emphatic particle, with a ratio of 28,554 to 38;
    \item \textcyrrm{da}: auxiliary particle (`to') vs. affirmative particle, with a ratio of 12,035 to 543;
    \item \textcyrrm{e}: 3rd person present auxiliary verb (`to be') vs. particle (`well') vs. interjection (`wow'), with a ratio of 9,136 to 21 to 5;
    \item singular masculine noun with a short definite article vs. count form of a masculine noun, with a ratio of 6,437 to 1,592;
    \item adverb vs. neuter singular adjective, with a ratio of 3,858 to 1,753.
\end{itemize}
\vspace{-1mm}

Overall,
the following factors should be taken into account when modeling Bulgarian morpho-syntax:
(1)~locality vs. non-locality of grammatical features,
(2)~interdependence of grammatical features, and
(3)~domain-specific preferences.

\section{Method}
\label{sec:method}

We used the guided learning framework described in \cite{shen-satta-joshi:2007:ACLMain},
which has yielded state-of-the-art results for English
and has been successfully applied to other morphologically complex languages
such as Icelandic \cite{dredze-wallenberg:2008:ACLShort};
we found it quite suitable for Bulgarian as well.
We used the feature set defined in \cite{shen-satta-joshi:2007:ACLMain},
which includes the following:

\vspace{-1mm}
\begin{enumerate}
\item The feature set of \newcite{Ratnaparkhi1996}, including prefix, suffix and
lexical, as well as some bigram and trigram context features;
\item Feature templates as in \cite{Ratnaparkhi1996}, which have been shown helpful in bidirectional search;
\item More bigram and trigram features and bi-lexical features as in \cite{shen-satta-joshi:2007:ACLMain}.
\end{enumerate}
\vspace{-1mm}

Note that we allowed prefixes and suffixes of length up to 9,
as in \cite{Toutanova:2003:FPT} and \cite{Tsuruoka:2005:BIE:1220575.1220634}.

We further extended the set of features with the tags proposed for the current word token by a morphological lexicon,
which maps words to possible tags;
it is exhaustive, i.e., the correct tag is always among the suggested ones for each token.

We also used 70 linguistically-motivated, high-precision rules
in order to further reduce the number of possible tags suggested by the lexicon.
The rules are similar to those proposed by \newcite{Hinrichs:2004} for German;
we implemented them as constraints in the CLaRK system~\cite{Simov:2003:CLaRK}.

Here is an example of a rule: If a wordform is ambiguous between
a masculine count noun (Ncmt) and a singular short definite
masculine noun (Ncmsh), the Ncmt tag should be chosen if the
previous token is a numeral or a number.

The 70 rules were developed by linguists based on observations over the training dataset only.
They target primarily the most frequent cases of ambiguity,
and to a lesser extent some infrequent but very problematic cases.
\joro{we shall provide an example for both?}
Some rules operate over classes of words, while other refer to particular wordforms.
The rules were designed to be 100\% accurate on our training dataset;
our experiments show that they are also 100\% accurate on the test and on the development dataset.

Note that some of the rules are dependent on others,
and thus the order of their cascaded application is important.
For example, the wordform \textcyrrm{ya} is ambiguous between
an accusative feminine singular short form of a personal pronoun (`her') and an interjection (`wow').
To handle this properly, the rule for interjection,
which targets sentence initial positions, followed by a comma, needs to be executed first.
The rule for personal pronouns is only applied afterwards.

\begin{table}[htb]
\centering
{\small \begin{tabular}{|r|l|l|}
\hline
\multicolumn{1}{|c|}{\bf Word} & \multicolumn{1}{c|}{\bf Tags}\\
\hline
\textcyrrm{To{\u i}} & \textbf{Ppe-os3m}\\
\textcyrrm{obache} & \textit{\scriptsize Cc}; \textbf{Dd}\\
\textcyrrm{nyama} & Afsi; Vnitf-o3s; Vnitf-r3s;\\
                  & Vpitf-o2s; Vpitf-o3s; \textbf{Vpitf-r3s}\\
\textcyrrm{v\cdprime zmozhnost} & \textbf{Ncfsi}\\
\textcyrrm{da} & \textit{\scriptsize Ta};\textbf{Tx}\\
\textcyrrm{sledi} & \textit{\scriptsize Ncfpi}; \textit{\scriptsize Vpitf-o2s}; \textit{\scriptsize Vpitf-o3s}; \textbf{Vpitf-r3s};\\
& \textit{\scriptsize Vpitz--2s}\\
$\ldots$ & $\ldots$\\
\hline
\end{tabular}}
\caption{
\label{tab:example:gazetteer:filtered}
\textbf{Sample fragment showing the possible tags suggested by the lexicon.}
        The tags that are further filtered by the rules are in italic;
        the correct tag is bold.}
\end{table}

The rules are quite efficient at reducing the POS ambiguity.
On the test dataset, before the rule application, 34.2\% of the tokens (excluding punctuation)
had more than one tag in our morphological lexicon.
This number is reduced to 18.5\% after the cascaded application of the 70 linguistic rules.
Table~\ref{tab:example:gazetteer:filtered} illustrates the effect of the rules on a small sentence fragment.
In this example, the rules have left only one tag (the correct one) for three of the ambiguous words.
Since the rules in essence decrease the average number of tags per token, we calculated that the lexicon
suggests 1.6 tags per token on average, and after the application of the rules this number decreases to 1.44 per token.

\preslav{Kakvo mozhe bi ni tryabva:
(1) Efekt na lexicon-a: (a) S kolko namalyava broyat na vyzmozhnite tagove; kolko taga na duma ostavat sredno. (b) Kolko pyti i procenti verniyat tag ne e sred tezi, koito sa predlozheni ot lexicon-a. There are no such situations.
(2) Efekt na lingvisti4nite pravila: (a) S kolko namalyavat tagovete. (b) Kolko pyti i procenti se premahva verniyat tag?}

\section{Datasets}
\label{sec:dataset}
                    
\subsection{BulTreeBank}

We used the latest version of the BulTreeBank~\cite{bul-morph:morpho},
which contains 20,556 sentences and 321,542 word tokens
(four times less than the English Penn Treebank),
annotated using a total of 680 unique morpho-syntactic tags.
See \cite{bul-morph} for a detailed description of the BulTreeBank tagset.

We split the data into training/development/test as shown in Table~\ref{tab:datasets}.
Note that only 552 of all 680 tag types were used in the training dataset,
and the development and the test datasets combined contain a total of 128 new tag types that were not seen in the training dataset.
Moreover, 32\% of the word types in the development dataset
and 31\% of those in the testing dataset do not occur in the training dataset.
Thus, data sparseness is an issue at two levels: word-level and tag-level.

\preslav{Hmm, kak li se sravnyava tova s angliyski ezik?}

\begin{table}[htb]
\centering
{\small \begin{tabular}{|l|r|r|r|c|}
\hline
\textbf{Dataset} & \textbf{Sentences} & \textbf{Tokens} & \textbf{Types} & \textbf{Tags}\\
\hline
Train & 16,532 & 253,526 & 38,659 & 552\\
Dev   &  2,007 &  32,995 &  9,635 & 425\\
Test  &  2,017 &  35,021 &  9,627 & 435\\
\hline
\end{tabular}}
\caption{
\label{tab:datasets}
\textbf{Statistics about our datasets.}}
\end{table}

\subsection{Morphological Lexicon}

In order to alleviate the data sparseness issues,
we further used a large morphological lexicon for Bulgarian,
which is an extended version of the dictionary described in \cite{PopovSimovVidinska98} and \cite{MorphoDict2003}.
It contains over 1.5M inflected wordforms (for 110K lemmata and 40K proper names),
each mapped to a set of possible morpho-syntactic tags.

\section{Experiments and Evaluation}
\label{sec:experiments}

State-of-the-art POS taggers for English
typically build a lexicon containing all tags a word type has taken in the training dataset;
this lexicon is then used to limit the set of possible tags that an input token can be assigned,
i.e., it imposes a hard constraint on the possibilities explored by the POS tagger.
For example, if \emph{can} has only been tagged as a verb and as a noun in the training dataset,
it will be only assigned those two tags at test time; other tags such as adjective, adverb and pronoun will not be considered.
Out-of-vocabulary words, i.e.,
those that were not seen in the training dataset,
are constrained as well, e.g., to a small set of frequent open-class tags.

In our experiments,
we used a morphological lexicon that is much larger than what could be built from the training corpus only:
building a lexicon from the training corpus only is of limited utility
since one can hardly expect to see in the training corpus all 52 synthetic forms a verb can possibly have.
Moreover, we did not use the tags listed in the lexicon as hard constraints (except in one of our baselines);
instead, we experimented with a different, non-restrictive approach:
we used the lexicon's predictions as features or soft constraints, i.e., as suggestions only,
thus allowing each token to take any possible tag.
Note that for both known and out-of-vocabulary words we used all 680 tags rather than the 552 tags observed in the training dataset;
we could afford to explore this huge search space thanks to the efficiency of the guided learning framework.
Allowing all 680 tags on training helped the model by exposing it to a larger set of negative examples.

We combined these lexicon features with standard features extracted from the training corpus.
We further experimented with the 70 contextual linguistic rules,
using them (a) as soft and (b) as hard constraints.
Finally, we set four baselines: three that do not use the lexicon and one that does.

\begin{table}[h]
\centering
{\small \begin{tabular}{|l|l|c|}
\hline
& & \textbf{Accuracy (\%)}\\
\textbf{\#} & \textbf{Baselines} & \textbf{(token-level)}\\
\hline
1 & MFT + unknowns are wrong & 78.10 \\
2 & MFT + unknowns are Ncmsi & 78.52 \\
3 & MFT + guesser for unknowns & 79.49 \\
\hline
4 & MFT + lexicon tag-classes & 94.40 \\
\hline
\end{tabular}}
\caption{
\label{tab:baselines}
\textbf{Most-frequent-tag (MFT) baselines.}}
\end{table}

\subsection{Baselines}

First, we experimented with the most-frequent-tag baseline,
which is standard for POS tagging.
This baseline ignores context altogether
and assigns each word type the POS tag it was most frequently seen with in the training dataset; ties are broken randomly.
We coped with word types not seen in the training dataset using three simple strategies:
(a)~we considered them all wrong,
(b)~we assigned them Ncmsi, which is the most frequent open-class tag in the training dataset, or
(c)~we used a very simple guesser,
which assigned Ncfsi, Ncnsi, Ncfsi, and Ncmsf,
if the target word ended by \textcyrrm{-a}, \textcyrrm{-o}, \textcyrrm{-i}, and \textcyrrm{-\cdprime t}, respectively,
otherwise, it assigned Ncmsi.
The results are shown in lines 1-3 of Table~\ref{tab:baselines}:
we can see that the token-level accuracy ranges in 78-80\% for (a)-(c),
which is relatively high, given that we use a large inventory of 680 morpho-syntactic tags.

We further tried a baseline that uses the above-described morphological lexicon,
in addition to the training dataset.
We first built two frequency lists, containing respectively
(1)~the most frequent tag in the training dataset for each word type, as before, and
(2)~the most frequent tag in the training dataset for each class of tags
that can be assigned to some word type, according to the lexicon.
For example, the most frequent tag for \textcyrrm{politika} is Ncfsi,
and the most frequent tag for the tag-class \{Ncmt;Ncmsi\} is Ncmt.

Given a target word type, this new baseline
first tries to assign it the most frequent tag from the first list.
If this is not possible, which happens (i)~in case of ties or (ii)~when the word type was not seen on training,
it extracts the tag-class from the lexicon and consults the second list.
If there is a single most frequent tag in the corpus for this tag-class, it is assigned;
otherwise a random tag from this tag-class is selected.

Line 4 of Table~\ref{tab:baselines} shows that this latter baseline achieves a very high accuracy of 94.40\%.
Note, however, that this is over-optimistic:
the lexicon contains a tag-class for each word type in our testing dataset, i.e.,
while there can be word types not seen in the training dataset,
there are no word types that are not listed in the lexicon.
Thus, this high accuracy is probably due to a large extent to the scale and quality of our morphological lexicon,
and it might not be as strong with smaller lexicons;
we plan to investigate this in future work.

\subsection{Lexicon Tags as Soft Constraints}


We experimented with three types of features:

\begin{enumerate}
\item Word-related features only;
\item Word-related features + the tags suggested by the lexicon;
\item Word-related features + the tags suggested by the lexicon but then further filtered using the 70 contextual linguistic rules.
\end{enumerate}

Table~\ref{tab:results} shows the sentence-level and the token-level accuracy on the test dataset
for the three kinds of features: shown on lines 1, 3 and 4, respectively.
We can see that using the tags proposed by the lexicon as features (lines 3 and 4) has a major positive impact,
yielding up to 49\% error reduction at the token-level and up to 37\% at the sentence-level,
as compared to using word-related features alone (line 1).

Interestingly, filtering the tags proposed by the lexicon using the 70 contextual linguistic rules
yields a minor decrease in accuracy both at the word token-level and at the sentence-level (compare line~4 to line~2).
This is surprising since the linguistic rules are extremely reliable:
they were designed to be 100\% accurate on the training dataset,
and we found them experimentally to be 100\% correct on the development and on the testing dataset as well.

One possible explanation is that by limiting the set of available tags for a given token at training time,
we prevent the model from observing some potentially useful negative examples.
We tested this hypothesis by using the unfiltered lexicon predictions at training time
but then making use of the filtered ones at testing time; the results are shown on line~5.
We can observe a small increase in accuracy compared to line~4:
from 97.80\% to 97.84\% at the token-level, and
from 70.30\% to 70.40\% at the sentence-level.
Although these differences are tiny,
they suggest that having more negative examples at training is helpful.

We can conclude that using the lexicon as a source of soft constraints
has a major positive impact,
e.g.,~because it provides access to important external knowledge
that is complementary to what can be learned from the training corpus alone;
the improvements when using linguistic rules as soft constraints are more limited.

\subsection{Linguistic Rules as Hard Constraints}


Next, we experimented with using the suggestions of the linguistic rules as hard constraints.
Table~\ref{tab:results} shows that this is a very good idea.
Comparing line 1 to line 2, which do not use the morphological lexicon, we can see very significant improvements:
from 95.72\% to 97.20\% at the token-level and from 52.95\% to 64.50\% at the sentence-level.
The improvements are smaller but still consistent when the morphological lexicon is used:
comparing lines 3 and 4 to lines 6 and 7, respectively,
we see an improvement from 97.83\% to 97.91\% and from 97.80\% to 97.93\% at the token-level,
and about 1\% absolute at the sentence-level.



\subsection{Increasing the Beam Size}

Finally, we increased the beam size of guided learning from 1 to 3 as in \cite{shen-satta-joshi:2007:ACLMain}.
Comparing line 7 to line 8 in Table~\ref{tab:results},
we can see that this yields further token-level improvement: from 97.93\% to 97.98\%.

\begin{table*}[htb]
\centering
{\small \begin{tabular}{|l|c|cc|c|cc|}
\hline
& \textbf{Lexicon} & \multicolumn{2}{c|}{\bf Linguistic Rules (applied to filter):} & \textbf{Beam} & \multicolumn{2}{c|}{\bf Accuracy (\%)}\\
\textbf{\#} & \textit{(source of)} & (a) \textit{the lexicon features} & (b) \textit{the output tags} & \textbf{size} & \textbf{Sentence-level} & \textbf{Token-level}\\
\hline
1 & -- & -- & -- & 1 & 52.95 & 95.72\\
2 & -- & -- & yes & 1 & 64.50  & 97.20\\
\hline
3 & features & -- & -- & 1 & 70.40  & 97.83\\
4 & features & yes & --& 1 & 70.30 & 97.80\\
5 & features & yes, \emph{for test only} & -- & 1 & 70.40 & 97.84\\
\hline
6 & features & -- & yes & 1 & 71.34 & 97.91\\
7 & features & yes & yes & 1 & 71.69 & 97.93\\
8 & features & yes & yes & 3 & 71.94 & 97.98\\
\hline
\end{tabular}}
\caption{
\label{tab:results}
\textbf{Evaluation results on the test dataset.}
Line 1 shows the evaluation results when using features derived from the text corpus only;
these features are used by all systems in the table.
Line 2 further uses the contextual linguistic rules to limit the set of possible POS tags that can be predicted.
Note that these rules (1) consult the lexicon, and (2) always predict a single POS tag.
Line 3 uses the POS tags listed in the lexicon as features, i.e., as soft suggestions only.
Line 4 is like line 3, but the list of feature-tags proposed by the lexicon is filtered by the contextual linguistic rules.
Line 5 is like line 4, but the linguistic rules filtering is only applied at test time; it is not done on training.
Lines 6 and 7 are similar to lines 3 and 4, respectively,
but here the linguistic rules are further applied to limit the set of possible POS tags that can be predicted,
i.e., the rules are used as hard constraints.
Finally, line 8 is like line 7, but here the beam size is increased to 3.}
\end{table*}

\preslav{Vpro4em, za kakvo izobshto polzvame dev set-a?}

\preslav{Hmm, ima li na4in da se sravni koe e po-dobre: tagovete ot lexicon-a da sa
(a) hard, t.e. da ograni4avat vyzmozhnite tagove, ili
(b) soft, t.e. da sa samo features, no klasifikatyoryt da mozhe da izbira absolyutno vsi4ki tagove za vseki token.}

\preslav{A kakvo pravim s nepoznatite dumi? Vsi4ki vyzmozhni tagove li im pozvolyavame? Mozhe bi tryabva da se ograni4at do open-class tags?}

\preslav{We need baselines: assign the most frequent tag from training to each known word; assign the most frequent tag overall to unknown words.}

\preslav{Maybe we need to report unknown word accuracy too}

\preslav{Kakyv e inter-annotator agreement za BulTreebank?}

\section{Discussion}
\label{sec:discussion}

\begin{table*}[htb]
\centering
{\small \begin{tabular}{|l|l|r|c|}
\hline
                 &                 &                  & \textbf{Accuracy}\\
\textbf{Tool/Authors} & \textbf{Method} & \textbf{\# Tags} & (token-level, \%)\\
\hline
*TreeTagger & Decision Trees & 680 & 89.21 \\
*ACOPOST & Memory-based Learning & 680 & 89.91 \\
*SVMtool & Support Vector Machines & 680 & 92.22 \\
*TnT & Hidden Markov Model & 680 & 92.53 \\
\hline
\cite{Georgiev:2009:CAB} & Maximum Entropy   & 680 & 90.34 \\
\cite{Simov:2001} & Recurrent Neural Network & 160 & 92.87 \\
\cite{Georgiev:2009:CAB} & Maximum Entropy   &  95 & 94.43 \\
\cite{Savkov:al:2011} & SVM + Lexicon + Rules & 680 & 94.65 \\
\cite{Tanev:Mitkov:2002} & Manual Rules & 303 & 95.00(=P=R) \\
\cite{Simov:2001} & Recurrent Neural Network & 15 & 95.17 \\
\cite{DBLP:conf/aimsa/DoychinovaM04} & Transformation-based Learning & 40 & 95.50 \\
\cite{DBLP:conf/aimsa/DoychinovaM04} & Manual Rules + Lexicon & 40 & 98.40 \\
\hline
                   & Guided Learning & 680 & 95.72 \\
                   & Guided Learning + Lexicon & 680 & 97.83 \\
\textbf{This work} & Guided Learning + Lexicon + Rules & 680 & 97.98 \\
\hline
                   & \textit{Guided Learning + Lexicon + Rules} & \textit{49} & \textit{98.85} \\
                   & \textit{Guided Learning + Lexicon + Rules} & \textit{13} & \textit{99.30} \\
\hline
\end{tabular}}
\caption{
\label{tab:comparison:bg}
\textbf{Comparison to previous work for Bulgarian.}
The first four lines report evaluation results for various standard POS tagging tools,
which were retrained and evaluated on the BulTreeBank.
The following lines report token-level accuracy for previously published work,
as compared to our own experiments using guided learning.}
\end{table*}

Table~\ref{tab:comparison:bg} compares our results to previously reported evaluation results for Bulgarian.
The first four lines show the token-level accuracy for standard POS tagging tools
trained and evaluated on the BulTreeBank:\footnote{We used the pre-trained TreeTagger;
for the rest, we report the accuracy given on the Webpage of the BulTreeBank:
\scriptsize{\texttt{www.bultreebank.org/taggers/taggers.html}}}
TreeTagger~\cite{Schmid:1994}, which uses decision trees,
TnT~\cite{Brants00tnt}, which uses a hidden Markov model,
SVMtool~\cite{Gimenez04}, which is based on support vector machines,
and ACOPOST~\cite{Ingo2002}, implementing the memory-based model of \newcite{Daelemans:al:1996}.
The following lines report the token-level accuracy reported in previous work,
as compared to our own experiments using guided learning.

We can see that we outperform by a very large margin (92.53\% vs. 97.98\%, which represents 73\% error reduction)
the systems from the first four lines,
which are directly comparable to our experiments:
they are trained and evaluated on the BulTreeBank using the full inventory of 680 tags.

We further achieved statistically significant improvement ($p < 0.0001$; Pearson's chi-squared test \cite{ChiSquare})
over the best pervious result on 680 tags:
from 94.65\% to 97.98\%,
which represents 62.24\% error reduction at the token-level.

Overall,
we improved over almost all previously published results.
Our accuracy is second only to the manual rules approach of \newcite{DBLP:conf/aimsa/DoychinovaM04}.
Note, however, that they used 40 tags only, i.e.,~their inventory is 17 times smaller than ours.
Moreover, they have optimized their tagset specifically to achieve very high POS tagging accuracy
by choosing not to attempt to resolve some inherently hard systematic ambiguities,
e.g.,~they do not try to choose between second and third person past singular verbs,
whose inflected forms are identical in Bulgarian and hard to distinguish when the subject is not present
(Bulgarian is a pro-drop language).

In order to compare our results more closely to the smaller tagsets in Table~\ref{tab:comparison:bg},
we evaluated our best model with respect to
(a)~the first letter of the tag only (which is part-of-speech only, no morphological information; 13 tags), e.g., Ncmsf becomes N,
and (b)~the first two letters of the tag (POS + limited morphological information; 49 tags), e.g., Ncmsf becomes Nc.
This yielded 99.30\% accuracy for (a) and 98.85\% for (b).
The latter improves over \cite{DBLP:conf/aimsa/DoychinovaM04}, while using a bit larger number of tags.

Our best token-level accuracy of 97.98\% is comparable and even slightly better
than the state-of-the-art results for English:
97.33\% when using Penn Treebank data only \cite{shen-satta-joshi:2007:ACLMain},
and 97.50\% for Penn Treebank plus some additional unlabeled data \cite{sogaard:2011:ACL-HLT20111}.
Of course, our results are only indirectly comparable to English.

Still, our performance is impressive because
(1)~our model is trained on 253,526 tokens only
while the standard training sections 0-18 of the Penn Treebank contain a total of 912,344 tokens, i.e.,~almost four times more,
and (2)~we predict 680 rather than just 48 tags as for the Penn Treebank, which is 14 times more.

Note, however, that
(1)~we used a large external morphological lexicon for Bulgarian, which yielded about 50\% error reduction
(without it, our accuracy was 95.72\% only),
and (2)~our train/dev/test sentences are generally shorter, and thus arguably simpler for a POS tagger to analyze:
we have 17.4 words per test sentence in the BulTreeBank vs. 23.7 in the Penn Treebank.

Our results also compare favorably to the state-of-the-art results for other morphologically complex languages that use large tagsets,
e.g., 95.2\% for Czech with 1,400+ tags \cite{hajickkop:2001},
92.1\% for Icelandic with 639 tags \cite{dredze-wallenberg:2008:ACLShort},
97.6\% for Arabic with 139 tags \cite{Habash:2005:ATP}.

\section{Error Analysis}
\label{sec:analysis}

In this section, we present error analysis
with respect to the impact of the POS tagger's performance on other processing steps in
a natural language processing pipeline, such as lemmatization and syntactic dependency parsing.

First, we explore the most frequently confused pairs of
tags for our best-performing POS tagging system;
these are shown in Table~\ref{tab:orth}.

We can see that most of the wrong tags share the same part-of-speech
(indicated by the initial uppercase letter), such as V for verb,
N for noun, etc. This means that most errors refer to the
morphosyntactic features. For example, personal or impersonal
verb; definite or indefinite feminine noun; singular or plural
masculine adjective, etc. At the same time, there are also cases,
where the error has to do with the part-of-speech label itself.
For example, between an adjective and an adverb,
or between a numeral and an indefinite pronoun.

\begin{table}[htb]
\centering
{\small \begin{tabular}{|r|l|l|}
\hline
\bf Freq. & \multicolumn{1}{c|}{\bf Gold Tag} & \bf Proposed Tag\\
\hline
43 & Ansi &  Dm \\
23 & Vpitf-r3s &  Vnitf-r3s \\
16 & Npmsh &  Npmsi \\
14 & Vpiif-r3s &  Vniif-r3s \\
13 & Npfsd &  Npfsi \\
12 & Dm &  Ansi \\
12 & Vpitcam-smi &  Vpitcao-smi \\
12 & Vpptf-r3p &  Vpitf-r3p \\
11 & Vpptf-r3s &  Vpptf-o3s \\
10 & Mcmsi &  Pfe-os-mi \\
10 & Ppetas3n &  Ppetas3m \\
10 & Ppetds3f &  Psot--3--f \\
9 & Npnsi &  Npnsd \\
9 & Vpptf-o3s &  Vpptf-r3s \\
8 & Dm &  A-pi \\
8 & Ppxts &  Ppxtd \\
7 & Mcfsi &  Pfe-os-fi \\
7 & Npfsi &  Npfsd \\
7 & Ppetas3m &  Ppetas3n \\
7 & Vnitf-r3s &  Vpitf-r3s \\
7 & Vpitcam-p-i &  Vpitcao-p-i \\
\hline
\end{tabular}}
\caption{
\label{tab:orth}
\textbf{Most frequently confused pairs of tags.}\\
}
\end{table}

We want to use the above tagger to develop
(1)~a rule-based lemmatizer, using the morphological lexicon, e.g., as in \cite{Plisson:al:2004}, and
(2)~a dependency parser like MaltParser
\cite{DBLP:journals/nle/NivreHNCEKMM07}, trained on the dependency part of the BulTreeBank.
We thus study the potential impact of wrong tags on the performance of these tools.

The lemmatizer relies on the lexicon and uses
string transformation functions defined via two operations -- {\em remove} and {\em concatenate}:

\vspace{6pt}
{\tt if} tag = Tag {\tt then}

\indent \indent \indent \{remove OldEnd; concatenate NewEnd\}
\vspace{6pt}

\noindent where Tag is the tag of the wordform, OldEnd is the
string that has to be removed from the end of the wordform, and
NewEnd is the string that has to be concatenated to the beginning
of the wordform in order to produce the lemma.

Here is an example of such a rule:

\vspace{3pt}
\indent \indent \indent {\tt if} tag = Vpitf-o1s {\tt then}

\indent \indent \indent \indent \indent \indent \{remove \textcyrrm{oh};
concatenate \textcyrrm{a}\}
\vspace{3pt}

The application of the above rule to the past simple verb form
\textcyrrm{chetoh} (`I read')
would remove \textcyrrm{oh}, and then concatenate \textcyrrm{a}.
The result would be the correct lemma \textcyrrm{cheta} (`to read').

Such rules are generated for each wordform in the morphological lexicon;
the above functional representation allows for compact
representation in a finite state automaton. Similar rules are
applied to the unknown words, where the lemmatizer tries to guess
the correct lemma.

Obviously, the applicability of each rule
crucially depends on the output of the POS tagger. If the tagger suggests the correct
tag, then the wordform would be lemmatized correctly. Note that,
in some cases of wrongly assigned POS tags in a given context, we
might still get the correct lemma. This is possible in the
majority of the erroneous cases in which the part-of-speech has been
assigned correctly, but the wrong grammatical alternative has
been selected. In such cases, the error does not influence lemmatization.

In order to calculate the proportion of such cases, we divided
each tag into two parts: (a)~grammatical features that are common
for all wordforms of a given lemma, and (b)~features that are
specific to the wordform.

The part-of-speech features are always
determined by the lemma.
For example, Bulgarian verbs have the lemma features {\em aspect} and {\em transitivity}.
If they are correct, then the lemma is
predicted also correctly, regardless of whether correct or wrong
on the grammatical features. For example, if the verb participle
form (aorist or imperfect) has its correct aspect and
transitivity, then it is lemmatized also correctly,
regardless of whether the imperfect or aorist features were guessed correctly;
similarly, for other error types. We evaluated these
cases for the 711 errors in our experiment, and we found that 206
of them (about 29\%) were non-problematic for lemmatization.

For the MaltParser, we encode most of the grammatical features of
the wordforms as specific features for the parser. Hence, it is
much harder to evaluate the problematic cases due to the tagger.
Still, we were able to make an estimation of some cases.
Our strategy was to ignore the grammatical features that do not
always contribute to the syntactic behavior of the wordforms.
Such grammatical features for the verbs are {\em aspect} and {\em
tense}. Thus, proposing perfective instead of imperfective
for a verb or present instead of past tense would not cause problems for the MaltParser.
Among our 711 errors, 190 cases (or about 27\%) were not problematic for parsing.

Finally, we should note that there are two special classes of
tokens for which it is generally hard to predict some of the
grammatical features: (1)~abbreviations and (2)~numerals written
with digits. In sentences, they participate in agreement
relations only if they are pronounced as whole phrases;
unfortunately, it is very hard for the tagger to guess such
relations since it does not have at its disposal enough features,
such as the inflection of the numeral form, that might help
detect and use the agreement pattern.

\section{Conclusion and Future Work}
\label{sec:conclusion}

We have presented experiments with part-of-speech tagging for Bulgarian,
a Slavic language with rich inflectional and derivational morphology.
Unlike most previous work for this language, which has limited the number of possible tags,
we used a very rich tagset of 680 morpho-syntactic tags as defined in the BulTreeBank.
By combining a large morphological lexicon with prior linguistic knowledge and guided learning
from a POS-annotated corpus, we achieved accuracy of 97.98\%,
which is a significant improvement over the state-of-the-art for Bulgarian.
Our token-level accuracy is also comparable to the best results reported for English.

In future work, we want to experiment with a richer set of features,
e.g.,~derived from unlabeled data \cite{sogaard:2011:ACL-HLT20111}
or from the Web \cite{Umansky-Pesin:2010:MWA,bansal-klein:2011:ACL-HLT2011}.
We further plan to explore ways to decompose the complex Bulgarian morpho-syntactic tags,
e.g.,~as proposed in \cite{Simov:2001} and \cite{smith-smith-tromble:2005:HLTEMNLP}.
Modeling long-distance syntactic dependencies \cite{dredze-wallenberg:2008:ACLShort}
is another promising direction;
we believe this can be implemented efficiently using posterior regularization \cite{graca:nips:2009}
or expectation constraints \cite{Bellare:2009:APL}.


\section*{Acknowledgments}
We would like to thank the anonymous reviewers for their useful comments,
which have helped us improve the paper.

The research presented above has been partially supported by the EU FP7 project 231720 EuroMatrixPlus,
and by the SmartBook project, funded by the Bulgarian National Science Fund under grant D002-111/15.12.2008.

\bibliographystyle{acl}
\bibliography{eacl2012}

\begin{thebibliography}{}

\bibitem[\protect\citename{Bansal and
  Klein}2011]{bansal-klein:2011:ACL-HLT2011}
Mohit Bansal and Dan Klein.
\newblock 2011.
\newblock Web-scale features for full-scale parsing.
\newblock In {\em Proceedings of the 49th Annual Meeting of the Association for
  Computational Linguistics: Human Language Technologies}, ACL-HLT '10, pages
  693--702, Portland, Oregon, USA.

\bibitem[\protect\citename{Bellare \bgroup et al.\egroup
  }2009]{Bellare:2009:APL}
Kedar Bellare, Gregory Druck, and Andrew McCallum.
\newblock 2009.
\newblock Alternating projections for learning with expectation constraints.
\newblock In {\em Proceedings of the 25th Conference on Uncertainty in
  Artificial Intelligence}, UAI '09, pages 43--50, Montreal, Quebec, Canada.

\bibitem[\protect\citename{Brants}2000]{Brants00tnt}
Thorsten Brants.
\newblock 2000.
\newblock {TnT} -- a statistical part-of-speech tagger.
\newblock In {\em Proceedings of the Sixth Applied Natural Language
  Processing}, ANLP '00, pages 224--231, Seattle, Washington, USA.

\bibitem[\protect\citename{Brill}1995]{Brill:1995:TEL}
Eric Brill.
\newblock 1995.
\newblock Transformation-based error-driven learning and natural language
  processing: a case study in part-of-speech tagging.
\newblock {\em Comput. Linguist.}, 21:543--565.

\bibitem[\protect\citename{Collins and Roark}2004]{collins-roark:2004:ACL}
Michael Collins and Brian Roark.
\newblock 2004.
\newblock Incremental parsing with the perceptron algorithm.
\newblock In {\em Proceedings of the 42nd Meeting of the Association for
  Computational Linguistics, Main Volume}, ACL '04, pages 111--118, Barcelona,
  Spain.

\bibitem[\protect\citename{Collins}2002]{Collins:2002:DTM}
Michael Collins.
\newblock 2002.
\newblock Discriminative training methods for hidden {M}arkov models: theory
  and experiments with perceptron algorithms.
\newblock In {\em Proceedings of the Conference on Empirical Methods in Natural
  Language Processing}, EMNLP '02, pages 1--8, Philadelphia, PA, USA.

\bibitem[\protect\citename{Daelemans \bgroup et al.\egroup
  }1996]{Daelemans:al:1996}
Walter Daelemans, Jakub Zavrel, Peter Berck, and Steven Gillis.
\newblock 1996.
\newblock {MBT}: A memory-based part of speech tagger generator.
\newblock In Eva Ejerhed and Ido Dagan, editors, {\em Fourth Workshop on Very
  Large Corpora}, pages 14--27, Copenhagen, Denmark.

\bibitem[\protect\citename{Dojchinova and
  Mihov}2004]{DBLP:conf/aimsa/DoychinovaM04}
Veselka Dojchinova and Stoyan Mihov.
\newblock 2004.
\newblock High performance part-of-speech tagging of {B}ulgarian.
\newblock In Christoph Bussler and Dieter Fensel, editors, {\em AIMSA}, volume
  3192 of {\em Lecture Notes in Computer Science}, pages 246--255. Springer.

\bibitem[\protect\citename{Dredze and
  Wallenberg}2008]{dredze-wallenberg:2008:ACLShort}
Mark Dredze and Joel Wallenberg.
\newblock 2008.
\newblock Icelandic data driven part of speech tagging.
\newblock In {\em Proceedings of the 44th Annual Meeting of the Association of
  Computational Linguistics: Short Papers}, ACL '08, pages 33--36, Columbus,
  Ohio, USA.

\bibitem[\protect\citename{Georgiev \bgroup et al.\egroup
  }2009]{Georgiev:2009:CAB}
Georgi Georgiev, Preslav Nakov, Petya Osenova, and Kiril Simov.
\newblock 2009.
\newblock Cross-lingual adaptation as a baseline: adapting maximum entropy
  models to {B}ulgarian.
\newblock In {\em Proceedings of the RANLP'09 Workshop on Adaptation of
  Language Resources and Technology to New Domains}, AdaptLRTtoND '09, pages
  35--38, Borovets, Bulgaria.

\bibitem[\protect\citename{Gim\'{e}nez and M\`{a}rquez}2004]{Gimenez04}
Jes\'{u}s Gim\'{e}nez and Llu\'{i}s M\`{a}rquez.
\newblock 2004.
\newblock {SVMTool}: A general {POS} tagger generator based on support vector
  machines.
\newblock In {\em Proceedings of the 4th International Conference on Language
  Resources and Evaluation}, LREC '04, Lisbon, Portugal.

\bibitem[\protect\citename{Graca \bgroup et al.\egroup }2009]{graca:nips:2009}
Joao Graca, Kuzman Ganchev, Ben Taskar, and Fernando Pereira.
\newblock 2009.
\newblock Posterior vs parameter sparsity in latent variable models.
\newblock In Yoshua Bengio, Dale Schuurmans, John~D. Lafferty, Christopher
  K.~I. Williams, and Aron Culotta, editors, {\em Advances in Neural
  Information Processing Systems 22}, NIPS '09, pages 664--672. Curran
  Associates, Inc., Vancouver, British Columbia, Canada.

\bibitem[\protect\citename{Habash and Rambow}2005]{Habash:2005:ATP}
Nizar Habash and Owen Rambow.
\newblock 2005.
\newblock Arabic tokenization, part-of-speech tagging and morphological
  disambiguation in one fell swoop.
\newblock In {\em Proceedings of the 43rd Annual Meeting of the Association for
  Computational Linguistics}, ACL '05, pages 573--580, Ann Arbor, Michigan.

\bibitem[\protect\citename{Haji\v{c} \bgroup et al.\egroup
  }2001]{hajickkop:2001}
Jan Haji\v{c}, Pavel Krbec, Pavel Kv\v{e}to\v{n}, Karel Oliva, and
  Vladim\'{\i}r Petkevi\v{c}.
\newblock 2001.
\newblock Serial combination of rules and statistics: A case study in {C}zech
  tagging.
\newblock In {\em Proceedings of the 39th Annual Meeting of the Association for
  Computational Linguistics}, ACL '01, pages 268--275, Toulouse, France.

\bibitem[\protect\citename{Haji\v{c}}1998]{hajic:1998}
Jan Haji\v{c}.
\newblock 1998.
\newblock {Building a Syntactically Annotated Corpus: The Prague Dependency
  Treebank}.
\newblock In Eva Haji\v{c}ov\'{a}, editor, {\em {Issues of Valency and Meaning.
  Studies in Honor of Jarmila Panevov\'{a}}}, pages 12--19. Prague Karolinum,
  Charles University Press.

\bibitem[\protect\citename{Hinrichs and Trushkina}2004]{Hinrichs:2004}
Erhard~W. Hinrichs and Julia~S. Trushkina.
\newblock 2004.
\newblock Forging agreement: Morphological disambiguation of noun phrases.
\newblock {\em Research on Language \& Computation}, 2:621--648.

\bibitem[\protect\citename{Johansson \bgroup et al.\egroup }1986]{LOB}
Stig Johansson, Eric Atwell, Roger Garside, and Geoffrey Leech, 1986.
\newblock {\em The Tagged {LOB} Corpus: Users' manual}.
\newblock ICAME, The Norwegian Computing Centre for the Humanities, Bergen
  University, Norway.

\bibitem[\protect\citename{Krushkov}1997]{Krushkov}
Hristo Krushkov.
\newblock 1997.
\newblock {\em Modelling and building machine dictionaries and morphological
  processors (in Bulgarian)}.
\newblock {Ph.D.} thesis, University of Plovdiv, Faculty of Mathematics and
  Informatics, Plovdiv, Bulgaria.

\bibitem[\protect\citename{Ku\v{c}era and
  Francis}1967]{kucera:computational:1967}
Henry Ku\v{c}era and Winthrop~Nelson Francis.
\newblock 1967.
\newblock {\em Computational analysis of present-day American English}.
\newblock Brown University Press, Providence, {RI}.

\bibitem[\protect\citename{Lafferty \bgroup et al.\egroup
  }2001]{Lafferty:2001:CRF}
John~D. Lafferty, Andrew McCallum, and Fernando C.~N. Pereira.
\newblock 2001.
\newblock Conditional random fields: Probabilistic models for segmenting and
  labeling sequence data.
\newblock In {\em Proceedings of the 18th International Conference on Machine
  Learning}, ICML '01, pages 282--289, San Francisco, CA, USA.

\bibitem[\protect\citename{Maamouri \bgroup et al.\egroup
  }2003]{Arabic:treebank}
Mohamed Maamouri, Ann Bies, Hubert Jin, and Tim Buckwalter.
\newblock 2003.
\newblock Arabic {T}reebank: Part 1 v 2.0.
\newblock LDC2003T06.

\bibitem[\protect\citename{Marcus \bgroup et al.\egroup
  }1993]{Marcus:1993:penntreebank}
Mitchell~P. Marcus, Mary~Ann Marcinkiewicz, and Beatrice Santorini.
\newblock 1993.
\newblock Building a large annotated corpus of {E}nglish: the {P}enn
  {T}reebank.
\newblock {\em Comput. Linguist.}, 19:313--330.

\bibitem[\protect\citename{Nivre \bgroup et al.\egroup
  }2007]{DBLP:journals/nle/NivreHNCEKMM07}
Joakim Nivre, Johan Hall, Jens Nilsson, Atanas Chanev, G{\"u}lsen Eryigit,
  Sandra K{\"u}bler, Svetoslav Marinov, and Erwin Marsi.
\newblock 2007.
\newblock {MaltParser}: A language-independent system for data-driven
  dependency parsing.
\newblock {\em Natural Language Engineering}, 13(2):95--135.

\bibitem[\protect\citename{Pind \bgroup et al.\egroup
  }1991]{Pind:Icelandic:dict}
J\"{o}rgen Pind, Fridrik Magn\'{u}sson, and Stef\'{a}n Briem.
\newblock 1991.
\newblock The {I}celandic frequency dictionary.
\newblock Technical report, The Institute of Lexicography, University of
  Iceland, Reykjavik, Iceland.

\bibitem[\protect\citename{Plackett}1983]{ChiSquare}
Robin~L. Plackett.
\newblock 1983.
\newblock {Karl Pearson and the Chi-Squared Test}.
\newblock {\em International Statistical Review / Revue Internationale de
  Statistique}, 51(1):59--72.

\bibitem[\protect\citename{Plisson \bgroup et al.\egroup
  }2004]{Plisson:al:2004}
Jo\"{e}l Plisson, Nada Lavra\v{c}, and Dunja Mladeni\'{c}.
\newblock 2004.
\newblock A rule based approach to word lemmatization.
\newblock In {\em Proceedings of the 7th International Multiconference:
  Information Society}, IS '2004, pages 83--86, Ljubljana, Slovenia.

\bibitem[\protect\citename{Popov \bgroup et al.\egroup
  }1998]{PopovSimovVidinska98}
Dimitar Popov, Kiril Simov, and Svetlomira Vidinska.
\newblock 1998.
\newblock {\em Dictionary of Writing, Pronunciation and Punctuation of
  Bulgarian Language (in Bulgarian)}.
\newblock Atlantis KL, Sofia, Bulgaria.

\bibitem[\protect\citename{Popov \bgroup et al.\egroup }2003]{MorphoDict2003}
Dimityr Popov, Kiril Simov, Svetlomira Vidinska, and Petya Osenova.
\newblock 2003.
\newblock {\em Spelling Dictionary of Bulgarian}.
\newblock Nauka i izkustvo, Sofia, Bulgaria.

\bibitem[\protect\citename{Ratnaparkhi}1996]{Ratnaparkhi1996}
Adwait Ratnaparkhi.
\newblock 1996.
\newblock A maximum entropy model for part-of-speech tagging.
\newblock In Eva Ejerhed and Ido Dagan, editors, {\em Fourth Workshop on Very
  Large Corpora}, pages 133--142, Copenhagen, Denmark.

\bibitem[\protect\citename{Savkov \bgroup et al.\egroup }2011]{Savkov:al:2011}
Aleksandar Savkov, Laska Laskova, Petya Osenova, Kiril Simov, and Stanislava
  Kancheva.
\newblock 2011.
\newblock A web-based morphological tagger for {B}ulgarian.
\newblock In Daniela Majchr\'{a}kov\'{a} and Radovan Garab\'{i}k, editors, {\em
  Slovko 2011. Sixth International Conference. Natural Language Processing,
  Multilinguality}, pages 126--137, Modra/Bratislava, Slovakia.

\bibitem[\protect\citename{Schmid}1994]{Schmid:1994}
Helmut Schmid.
\newblock 1994.
\newblock Probabilistic part-of-speech tagging using decision trees.
\newblock In {\em International Conference on New Methods in Language
  Processing}, pages 44--49, Manchester, UK.

\bibitem[\protect\citename{Schr\"{o}der}2002]{Ingo2002}
Ingo Schr\"{o}der.
\newblock 2002.
\newblock A case study in part-of-speech-tagging using the {ICOPOST} toolkit.
\newblock Technical Report FBI-HH-M-314/02, Department of Computer Science,
  University of Hamburg.

\bibitem[\protect\citename{Shen \bgroup et al.\egroup
  }2007]{shen-satta-joshi:2007:ACLMain}
Libin Shen, Giorgio Satta, and Aravind Joshi.
\newblock 2007.
\newblock Guided learning for bidirectional sequence classification.
\newblock In {\em Proceedings of the 45th Annual Meeting of the Association of
  Computational Linguistics}, ACL '07, pages 760--767, Prague, Czech Republic.

\bibitem[\protect\citename{Simov and Osenova}2001]{Simov:2001}
Kiril Simov and Petya Osenova.
\newblock 2001.
\newblock A hybrid system for morphosyntactic disambiguation in {B}ulgarian.
\newblock In {\em Proceedings of the EuroConference on Recent Advances in
  Natural Language Processing}, RANLP '01, pages 5--7, Tzigov chark, Bulgaria.

\bibitem[\protect\citename{Simov and Osenova}2004]{bul-morph:morpho}
Kiril Simov and Petya Osenova.
\newblock 2004.
\newblock {BTB-TR04}: {BulTreeBank} morphosyntactic annotation of {B}ulgarian
  texts.
\newblock Technical Report BTB-TR04, Bulgarian Academy of Sciences.

\bibitem[\protect\citename{Simov \bgroup et al.\egroup }2003]{Simov:2003:CLaRK}
Kiril~Ivanov Simov, Alexander Simov, Milen Kouylekov, Krasimira Ivanova, Ilko
  Grigorov, and Hristo Ganev.
\newblock 2003.
\newblock Development of corpora within the {CLaRK} system: The {BulTreeBank}
  project experience.
\newblock In {\em Proceedings of the 10th conference of the European chapter of
  the Association for Computational Linguistics}, EACL '03, pages 243--246,
  Budapest, Hungary.

\bibitem[\protect\citename{Simov \bgroup et al.\egroup }2004]{bul-morph}
Kiril Simov, Petya Osenova, and Milena Slavcheva.
\newblock 2004.
\newblock {BTB-TR03}: {BulTreeBank} morphosyntactic tagset.
\newblock Technical Report BTB-TR03, Bulgarian Academy of Sciences.

\bibitem[\protect\citename{Smith \bgroup et al.\egroup
  }2005]{smith-smith-tromble:2005:HLTEMNLP}
Noah~A. Smith, David~A. Smith, and Roy~W. Tromble.
\newblock 2005.
\newblock Context-based morphological disambiguation with random fields.
\newblock In {\em Proceedings of Human Language Technology Conference and
  Conference on Empirical Methods in Natural Language Processing}, pages
  475--482, Vancouver, British Columbia, Canada.

\bibitem[\protect\citename{S{\o}gaard}2011]{sogaard:2011:ACL-HLT20111}
Anders S{\o}gaard.
\newblock 2011.
\newblock Semi-supervised condensed nearest neighbor for part-of-speech
  tagging.
\newblock In {\em Proceedings of the 49th Annual Meeting of the Association for
  Computational Linguistics}, ACL-HLT '10, pages 48--52, Portland, Oregon, USA.

\bibitem[\protect\citename{Tanev and Mitkov}2002]{Tanev:Mitkov:2002}
Hristo Tanev and Ruslan Mitkov.
\newblock 2002.
\newblock Shallow language processing architecture for {B}ulgarian.
\newblock In {\em Proceedings of the 19th International Conference on
  Computational Linguistics}, COLING '02, pages 1--7, Taipei, Taiwan.

\bibitem[\protect\citename{Toutanova \bgroup et al.\egroup
  }2003]{Toutanova:2003:FPT}
Kristina Toutanova, Dan Klein, Christopher~D. Manning, and Yoram Singer.
\newblock 2003.
\newblock Feature-rich part-of-speech tagging with a cyclic dependency network.
\newblock In {\em Proceedings of the Conference of the North American Chapter
  of the Association for Computational Linguistics}, NAACL '03, pages 173--180,
  Edmonton, Canada.

\bibitem[\protect\citename{Tsuruoka and
  Tsujii}2005]{Tsuruoka:2005:BIE:1220575.1220634}
Yoshimasa Tsuruoka and Jun'ichi Tsujii.
\newblock 2005.
\newblock Bidirectional inference with the easiest-first strategy for tagging
  sequence data.
\newblock In {\em Proceedings of the Conference on Human Language Technology
  and Empirical Methods in Natural Language Processing}, HLT-EMNLP '05, pages
  467--474, Vancouver, British Columbia, Canada.

\bibitem[\protect\citename{Tsuruoka \bgroup et al.\egroup
  }2011]{tsuruoka-miyao-kazama:2011:CoNLL}
Yoshimasa Tsuruoka, Yusuke Miyao, and Jun'ichi Kazama.
\newblock 2011.
\newblock Learning with lookahead: Can history-based models rival globally
  optimized models?
\newblock In {\em Proceedings of the 49th Annual Meeting of the Association for
  Computational Linguistics: Human Language Technologies}, ACL-HLT '10, pages
  238--246, Portland, Oregon, USA.

\bibitem[\protect\citename{Umansky-Pesin \bgroup et al.\egroup
  }2010]{Umansky-Pesin:2010:MWA}
Shulamit Umansky-Pesin, Roi Reichart, and Ari Rappoport.
\newblock 2010.
\newblock A multi-domain web-based algorithm for {POS} tagging of unknown
  words.
\newblock In {\em Proceedings of the 23rd International Conference on
  Computational Linguistics: Posters}, COLING '10, pages 1274--1282, Beijing,
  China.

\end{thebibliography}

\end{document}